\pdfoutput=1

\documentclass[11pt]{article}

\usepackage{acl}

\usepackage{times}
\usepackage{latexsym}

\usepackage[T1]{fontenc}

\usepackage[utf8]{inputenc}

\usepackage{microtype}

\usepackage{tcolorbox}
\usepackage{arydshln}
\usepackage{algorithm}
\usepackage{algpseudocode}
\usepackage{ulem}
\usepackage{soul}
\usepackage{xcolor}
\usepackage{todonotes}
\definecolor{mybluei}{RGB}{0,173,239}

\usepackage{xspace}
\newcommand{\method}{\mbox{\textsc{TRACER}}\xspace}

\algnewcommand{\LeftComment}[1]{\Statex \(\triangleright\) #1}

\makeatletter
\def\els@aparagraph[#1]#2{\elsparagraph[#1]{#2}}
\def\els@bparagraph#1{\elsparagraph*{#1}}
\makeatother

\title{Wizard of Shopping: Target-Oriented E-commerce Dialogue Generation with Decision Tree Branching}

\author{Xiangci Li\textsuperscript{\rm 1}$^*$ ~~ Zhiyu Chen\textsuperscript{\rm 2} ~~ Jason Ingyu Choi\textsuperscript{\rm 2}
\AND Nikhita Vedula\textsuperscript{\rm 2} ~~ Besnik Fetahu\textsuperscript{\rm 2} ~~ Oleg Rokhlenko\textsuperscript{\rm 2} ~~ Shervin Malmasi\textsuperscript{\rm 2} \\
  \textsuperscript{\rm 1} AWS AI Labs 
  \textsuperscript{\rm 2} Amazon.com, Inc. \\
  \tt lixiangci8@gmail.com \\ \tt \{zhiyuche, chojson, veduln, besnikf, olegro, malmasi\}@amazon.com \\
}

\begin{document}
\maketitle

\def\thefootnote{*}\footnotetext{~Work performed by the author as a PhD candidate at The University of Texas at Dallas before joining AWS AI Labs.}
\def\thefootnote{\arabic{footnote}}

\begin{abstract}
The goal of conversational product search (CPS) is to develop an intelligent, chat-based shopping assistant that can directly interact with customers to understand shopping intents, ask clarification questions, and find relevant products. However, training such assistants is hindered mainly due to the lack of reliable and large-scale datasets. 
Prior human-annotated CPS datasets are extremely small in size and lack integration with real-world product search systems. 
We propose a novel approach, \method, which leverages large language models (LLMs) to generate realistic and natural conversations for different shopping domains. \method's novelty lies in grounding the generation to dialogue plans, which are product search trajectories predicted from a decision tree model, that guarantees relevant product discovery in the shortest number of search conditions. 
We also release the first target-oriented CPS dataset \textit{Wizard of Shopping (WoS)}, containing highly natural and coherent conversations (3.6k) from three shopping domains. Finally, we demonstrate the quality and effectiveness of \textit{WoS} via human evaluations and downstream tasks.

\end{abstract}

\section{Introduction}\label{sec:intro}

E-commerce platforms use product search to help customers find items in billion-scale product catalogs. 
Current product search systems rely on a \textit{search-and-refine} process, where customers iteratively refine keywords expressing their shopping needs and then assess the top--$k$ results for relevance. 
This process is customer-driven, and without sufficient expertise or patience, may lead to suboptimal purchase decisions or shopping journey abandonment. 
Conversational Product Search (CPS)~\cite{zhang2018towards,bi2019conversational,zou2022learning} 
address the limitations of traditional product search by adding a conversational layer into product search, which provides a more natural and delightful shopping experience than traditional systems.


\begin{figure}[t!]
\centering
\includegraphics[width=0.5\textwidth]{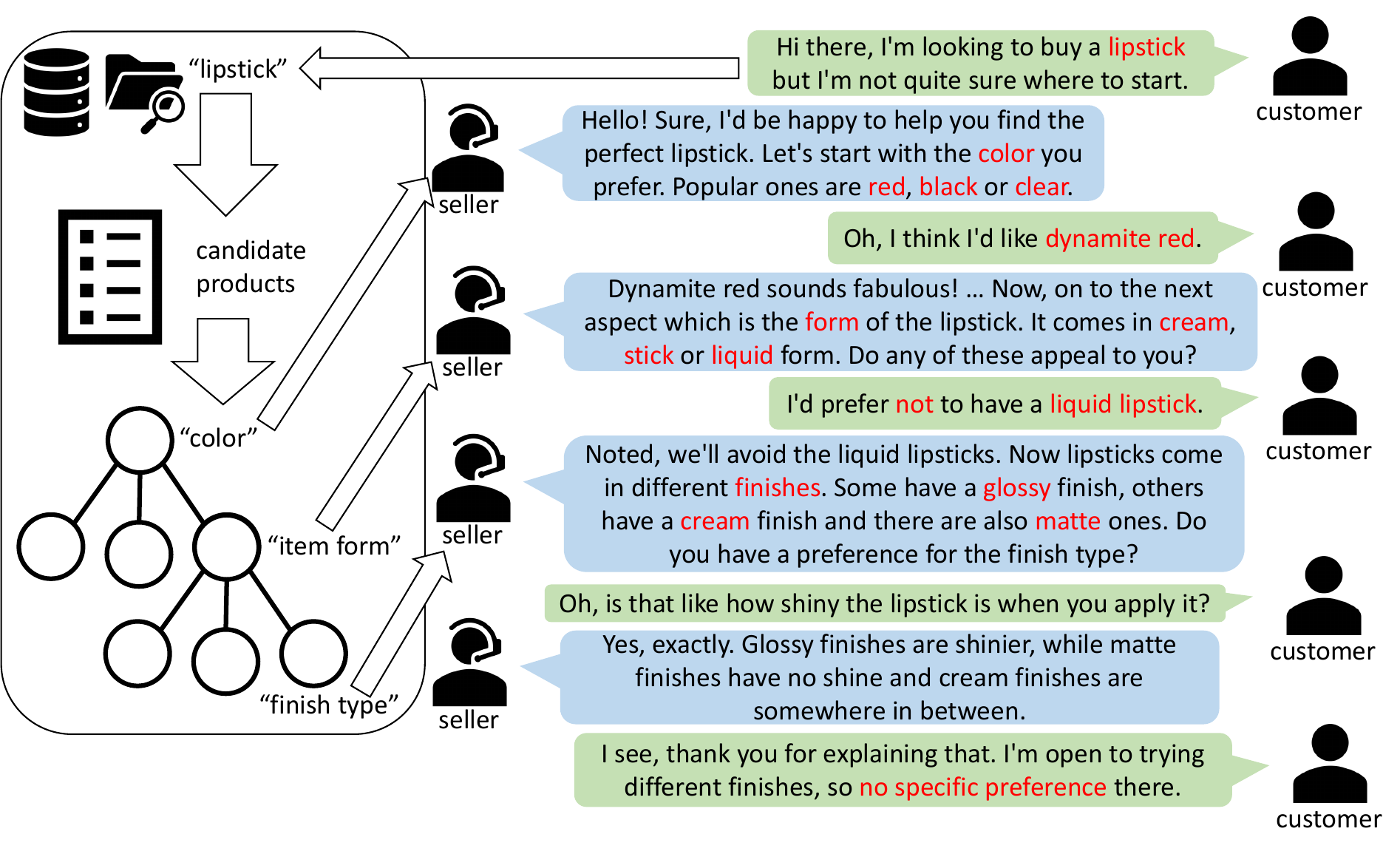}
\vspace{-2.5em}
\caption{An example of generated shopping conversation using \method, which leverages the LLM with different roles and decision tree-based planning.
}
\vspace{-1.5em}
\label{fig:illustration}
\end{figure}


%

Nonetheless, existing product search systems are not optimized for either natural language queries or real dialogues. One solution involves training
a model to rewrite conversations into decontextualized queries \cite{yu2020few, wu2022conqrr}. 
However, prior rewriting datasets do not cover the shopping domain. 
Another solution is to directly train CPS models with dialogues 
~\cite{zhang2018towards,bi2019conversational,zou2022learning}, but this approach often relies on synthetic datasets with templated content, which limits the realism of the trained models. Consequently, the effectiveness of these models on real dialogues remains uncertain. Recently, \citet{Bernard:2023:SIGIR} released a small-scale dataset with expensive human annotation. 
Overall, the scarcity of datasets presents a significant challenge in developing intelligent CPS assistants.

In order to tackle the above challenge, we propose a novel approach, \method (\textbf{T}arget-oriented e-comme\textbf{R}ce di\textbf{A}logue generation with de\textbf{C}ision tr\textbf{E}e b\textbf{R}anching) to simulate shopping conversations between two LLM agents. Specifically, we assign two distinct roles to each LLM: (1) a \underline{customer} with specified shopping interests (a.k.a. the "Apprentice"); 
(2) a \underline{seller} agent (a.k.a. the "Wizard") that has access to a product catalog. 
%
To overcome the known shortcomings of LLMs, such as hallucination, prompt brittleness, lack of controllability, and planning~\cite{kaddour2023challenges}, 
two agents leverage dialogue plans predicted from decision tree to ground dialogue generation. Since the decision tree is trained to select product attributes that maximally split the search space, customers are guaranteed to reach target products in the shortest number of search conditions, minimizing user effort \cite{al2010review}. Human evaluations further validate that the generated dialogues are highly natural and coherent since LLMs can easily convert dialogue plans to natural conversations (See Figure \ref{fig:illustration} as an example). Our contributions are the following:
\vspace{-0.5em}
\begin{itemize}
\setlength\itemsep{-0.5em}
    \item A highly controllable, scalable, and easy-to-apply dialog generation approach \method for different shopping domains.
    \item Release of \textit{Wizard-of-Shopping} (\textit{WoS}) dataset which includes 3.6k shopping conversations across 3 different product domains.
    \item Demonstrate the utility of \textit{WoS} by showing improvements in conversational query generation and product ranking tasks.
\end{itemize}

\section{Related Work} \label{sec:related_work}

\vspace{-0.5em}
\paragraph{Conversational Product Search.}

\citet{zhang2018towards} proposed a seminal CPS system that unified conversational product search and recommendation, and actively asked questions to understand user needs. 
Their framework predicts both the true item and the next question at each conversation turn. 
\citet{bi2019conversational} extended this by predicting the probability of an aspect-value pair being positive or negative to enhance conversational product ranking performance. 
\citet{zou2022learning} learned a joint representation of users, queries, items, and conversations to retrieve target items within a latent semantic space. 
\citet{xiao-etal-2021-end} created an online shopping conversational dataset by transferring the utterance structures from movie domain conversations. They used proprietary search behavior data to supervise the construction of intent flow.

A fundamental issue in prior studies~\cite{zhang2018towards,bi2019conversational,zou2022learning} is that all the proposed CPS models are evaluated using slotted templates specifically designed for clarification questions. 
The weaknesses of such an evaluation are twofold. First, as the quality of simulated conversations is not assessed and may be of poor quality, the trained conversational product rankers might not generalize well to real conversations. Second, customers may lack sufficient product knowledge, making it unrealistic to expect them to always provide answers to the seller's questions. 

The paucity of datasets significantly hinders CPS research. Many datasets used in prior works \cite{xiao-etal-2021-end, chen-etal-2020-jddc, zhao-etal-2022-jddc} are privately owned, which prevents reproduction and further reusing of their work for future studies.
On the other hand, \citet{Bernard:2023:SIGIR} released MG-ShopDial, a 64-conversation human-generated e-commerce dialogue dataset, 
we perform an in-depth comparison in \S\ref{sec:mgshopdial_comparison}. Moreover, \textit{due to commercial interests and privacy concerns, no large-scale human-generated CPS dataset is publicly available, and it will not likely exist in the future either}. 

\textit{Our work aims to address the data scarcity issue by guiding LLMs to generate natural CPS conversations.} We prompt LLMs to generate conversations grounded on sampled customer preferences and product information from a real product catalog. 
The customer is also allowed to ask sellers general knowledge questions, 
which is lacking in prior work~\cite{zhang2018towards,bi2019conversational,zou2022learning}. Overall, our generated dialogues in the \textit{WoS} dataset are more natural and diverse than prior simulated conversations. 
We believe that our proposed method and datasets will be a valuable contribution to CPS and its sub-tasks.

\vspace{-0.5em}
\paragraph{Difference with Prior Work in Conversational Recommendation \& Search.}
Similar to previous CPS work, prior conversational recommendation (CR) systems studies \cite{li2018towards, hayati2020inspired, liu2020towards, liu2021durecdial, deng2021unified, zhang2022multiple} explored the recommendation strategies by learning both the user and item representations. Indeed, the problem of learning a conversation strategy for the seller to find proper product aspects that shorten the product-seeking process seems to be shared between CPS and CR.
A key difference between our work and prior studies is that \textit{we do not build a new CPS or CR system}, but instead focus on the creation of a dataset with natural conversations along with dialogue plans with decision trees and LLMs so that the dataset can be applied to improve the downstream tasks such as query generation and product ranking. Notably, our product retrievals build upon traditional approaches like BM25, incorporating conversational layers to achieve CPS. This sets our work apart from previous CPS efforts~\cite{zhang2018towards,bi2019conversational,zou2022learning}. Additionally, many CR works are in the movie domain \cite{li2018towards, hayati2020inspired, liu2020towards, liu2021durecdial} and there is no existing CPS dataset specifically tailored for the e-commerce domain. Our work addresses this gap by introducing a dataset, along with its corresponding dataset generation method, designed explicitly for the e-commerce domain.

\section{Dialogue Generation Approach} 
\label{sec:approach}
As shown in Figure \ref{fig:illustration}, the goal is to generate a shopping dialogue given a selected set of preferences that is derived from a product ($p$), randomly sampled from a product catalog. \method generates dialogues in three steps:

1. Customer preference sampling (\S\ref{sec:preference_sampling}) assigns relevant product attributes from $p$ into three distinct preference groups. 

2. Dialogue planning (\S\ref{sec:overall_dialogue_generation}) fits decision trees to predict the sequence of product aspects to be addressed so that customers can find desired products with minimal effort. 

3. Verbalization (\S\ref{sec:verbalization}) utilizes customer preference, planned search trajectories, and prompt engineering to generate natural and relevant dialogues.



\begin{algorithm}[H]
\small
\caption{Decision Tree-Based Search Strategy}\label{alg:decision_tree_search}
\begin{algorithmic}
\Function{DecisionTreeSearch}{$RevPref$}
\State $P_o \gets search(RevPref)$
\State $X, Y \gets MakeDataset(P_o)$
\State $Tree \gets DecisionTree.fit(X,Y)$
\State \Return \textit{$P_o, Tree$}
\EndFunction
\end{algorithmic}
\end{algorithm}
\vspace{-1.5em}

\begin{algorithm}[t!]

\small
\caption{Interactive and single-pass dialogue generation strategies from \S\ref{sec:verbalization}.}
\label{alg:decision_tree_dialogue_generation}
\begin{algorithmic}
\Require $LLM$, $preference$ 


\LeftComment{\textbf{Stage 1}: Customer starts with a product category.}
    \State $Conv\_hist \gets []$ \Comment{Interactive generation}
    \State $Utterances \gets LLM.verbalize\_stage1(PC)$
    \State $Conv\_hist.extend(Utterances)$
\LeftComment{\textbf{Stage 2}: Product search.}
\State $Plan, Plan_{hist} \gets [], []$
\While{$|P_o|$ has not converged}
\State $RevPref \gets [PC, Plan_{hist}]$
\State $P_o, Tree \gets DecisionTreeSearch(RevPref)$
\State $Plan \gets Tree.traverse(preference)$ 
\LeftComment{$Plan = \{(A_1, V_1, I_1), ..., (A_d, V_d, I_d)\}$}
\State $Plan_{hist}.extend(Plan)$
    \State $Utterances \gets LLM.verbalize\_stage2(Plan, Conv\_hist)$
    \State $Conv\_hist.extend(Utterances)$
\EndWhile

\LeftComment{\textbf{Stage 3}: Product recommendation.}
\State $p \gets RandomSample(P_o)$
\State $Utterances \gets LLM.verbalize\_stage3(p, Conv\_hist)$ 
\State $Conv\_hist.extend(Utterances)$
\LeftComment{Single pass generation}
\State $Single\_pass\_conv \gets LLM.verbalize(preference, Plan_{hist}, p)$ 

\State \textbf{Output}  $Conv\_hist$, $Single\_pass\_conv$
\end{algorithmic}
\end{algorithm}
\vspace{-0.5em}

\subsection{Customer Preference Sampling} \label{sec:preference_sampling}

Each product is associated with a varying number of features or aspect-value pairs, e.g. $\{(A_1, V_1), (A_2, V_2), ..., (A_m, V_m)\}$. 
For example, a tablet case can have associated features of \{(model, iPad), (color, blue), (material, TPU)\}. 
We randomly assign one of the following customer interest values ($I$) to each aspect of the sampled product $p$ :
\vspace{-1em}
\begin{itemize}
\setlength\itemsep{-0.5em}
    \item \textbf{Wanted}: customer is interested
    \item \textbf{Unwanted}: customer is NOT interested
    \item \textbf{Optional}: customer does not care much
\end{itemize}
\vspace{-1em}

The \textit{wanted} aspect-value pairs come directly from $p$. As for an aspect-value pair $(A_x,V_x)$ from $p$ assigned with \textit{unwanted}, we randomly replace $V_x$ with a value $V'_x \ne V_x$ as the customer's unpreferred value, where $V'_x$ appears as a value for $A_x$ in the product catalog, so that the \textit{unwanted} pair becomes $(A_x,V'_x)$. 
For example, if $A_x$=``color'' and $V_x$=``blue'', $V'_x$ can be ``red''. 
We use the Product Category ($PC$) of product $p$, as well as the sampled \textit{wanted}, \textit{optional}, and \textit{unwanted} features to simulate the \textit{preference} of the customer agent: 
\vspace{-0.8em}
\begin{equation}
\small
    preference = [PC, (A_1,V_1,I_1), ..., (A_m,V_m,I_m)]
\vspace{-1em}
\end{equation}

where $I_i \in (wanted, unwanted, optional)$. When $I_i=optional$, we set $V_i$ as empty since the customer does not care about the value of the optional aspect.

\begin{table}
\small
\begin{center}
    \begin{tabular}{p{0.95\linewidth} }
    \hline
    \textbf{Instruction} \\ 
    You are a scriptwriter. For the aspects or aspect value pairs below, write a chat conversation between the customer and seller about the seller trying to narrow down the customer's need. \textbf{\textit{The seller knows well about the sold products, while the customer has limited knowledge about the products.}} [...] The customer should speak out first and say he/she wants to buy \{\{\textit{ProductCategory}\}\}. [...] You MUST cover ALL aspect-value pairs below, including the wanted, unwanted, and optional aspect-value pairs. \textbf{\textit{Make sure you reorganize and reorder the aspect value pairs and translate them into a natural conversation in a meaningful way.}} [...]  \\ \hdashline
    \textbf{Wanted / Optional / Unwanted Features} \\ 
  Aspect or aspect value pairs that the customer must say \{he wants to have / is optional / he does not want to have\} for the target product:  \\ 
  Aspect: \{\{\textit{Aspect}\}\}, Value: \{\{\textit{Value}\}\}; \\ 
  ... \\\hdashline
  \textbf{Candidate Value Examples} \\ 
    \textbf{\textit{Additionally, there are some common values for each aspect to be mentioned. The seller should use these typical values in his question:}} \\
    Aspect: \{\{\textit{Aspect}\}\}, Typical Values: \{\{\textit{$Value_1$}\}\}, \{\{\textit{$Value_2$}\}\}, \{\{\textit{$Value_3$}\}\}; \\ 
  ... \\\hdashline
\textbf{Final Instruction} \\
Output a conversation following all instructions above and make sure you double check if ALL aspect value pairs above are mentioned: \\ \hline
    \end{tabular}
    \caption{Simplified prompt format for single pass dialogue generation approach. Prompts in bold are designed for improving conversation naturalness (\S\ref{sec:improving_naturalness}).}
    \label{tab:simple_prompt}
    \vspace{-1.5em}
\end{center}
\end{table}

\subsection{Dialogue Planning with Decision Tree} \label{sec:overall_dialogue_generation}
Given a sampled \textit{preference}, dialogue planning as the next step determines the sequence of product aspects to be addressed in the conversation. 
Since previous studies on web search evaluation discovered that user effort is inversely correlated with user satisfaction \cite{al2010review}, we claim that real customers will also want to minimize their efforts in CPS. To reflect this intuition in dialogue generation, we focus on controlling the ``customer'' LLM to be aligned with expected customer behavior by intelligently deciding what to ask at each turn.

An intuitive approach is to utilize the concept of purity in decision trees \cite{mehta1996sliq, quinlan2014c4}. At each turn, given \textit{partially} revealed preference $RevPref$\footnote{$RevPref$ is initialized with $PC$ only and grows as more product features are revealed by the decision trees.}, the decision tree selects the next product attribute that maximally divides the current search space. This is important because in real applications, customers are also expected to reveal their preferences step-by-step. 

As Algorithm \ref{alg:decision_tree_search} shows, first, at each turn that requires search, a set of products ($P_o$) that satisfy $RevPref$ are retrieved from our search system. Note that our search system is a simple rule-based system that can filter out products based on $RevPref$. Then, product attribute features from $P_o$ are used for constructing a temporary training dataset to fit a decision tree, and we use all aspects from product catalog as features for each product. If a product does not have certain aspects, corresponding values are treated as empty. For example, let's consider a product from the \textit{tablet case} category with three features. Input features ($X$) can be represented as \{(model, iPad), (color, blue), (material, TPU)\} and the corresponding labels ($Y$) are represented as a string ``\textit{model:iPad\&color:blue\&material:TPU}''. In other words, products associated with the same aspect-value pairs all belong to the same leaf node of the decision tree. As a result, the fitted $Tree$ contains a structured collection of nodes whose attributes are the product aspect keys and split conditions are corresponding aspect values. 



Lastly, given the fitted $Tree$ and $preference$, the best dialogue plan (a sequence of aspect-value-interest tuples) is generated. Specifically, we traverse starting from the root and choose the next child based on the customer $preference$'s corresponding values and interest of the product aspect of the parent node. Any traversed aspect that does not appear in $preference$ is $optional$, indicating the customer does not care about the product aspect. Note that decision tree is fitted individually for each aspect selection step and used together with additional strategies~
(\S \ref{sec:improving_naturalness}) to increase the naturalness of generated dialogues. This search and planning process (Algorithm \ref{alg:decision_tree_search}) iterates until every product in $P_0$ satisfies $preference$, or until there are no more product aspects to discuss (Algorithm \ref{alg:decision_tree_dialogue_generation} Stage 2).

\subsection{Verbalization} 
\label{sec:verbalization}
We feed the given $preference$ and dialogue plan contexts through zero-shot prompting to generate dialogues, proposing two methods of verbalization. 


\vspace{-0.5em}
\paragraph{Interactive Generation.}
\label{sec:IG}
We use different prompts for the seller and customer (Appendix Tables \ref{tab:seller_prompt} and \ref{tab:customer_prompt}) agents so that they alternatively speak to each other utterance-by-utterance, like in the real world. 
The agents respond based on the previous conversation history. To ensure that the LLM-powered agents follow the instructions to adhere to the pre-determined features, we use a \textit{dialogue state tracker} \cite{young2010hidden} to track the customer's \textit{wanted}, \textit{optional}, and \textit{unwanted} features, that have been mentioned and are to be mentioned after each utterance. This is implemented by a hybrid strategy of rule-based keyword matching and a GPT-4 functional call prompt similar to prompts in Table \ref{tab:customer_prompt}, except that the expected output is the updated inputs to the ``update\_dialogue\_state'' function in Table \ref{tab:customer_prompt} instead of response utterances. This complex mechanism implies that the interactive approach is more prone to errors, as we show in \S\ref{sec:human_evaluation}.

\vspace{-0.5em}
\paragraph{Single-Pass Generation.}
\label{sec:SG}
Alternatively, we propose a simpler approach by feeding all product features pre-determined by the decision tree plan into a single LLM, and generating the entire conversation with a single pass (Table \ref{tab:simple_prompt}, see full prompts in Table \ref{tab:single_pass_prompt}). Despite we lose the interactive capacity, the generated conversations have fewer errors, as we show in \S\ref{sec:human_evaluation}.

\subsection{Enhancing Conversational Naturalness}
Our overall approach is described in Algorithm \ref{alg:decision_tree_dialogue_generation}.
\vspace{-0.5em}
\paragraph{How conversation starts.}
When dialogue starts, similar to prior work~\cite{zhang2018towards,bi2019conversational,zou2022learning}, each customer always starts with the target product category as an initial request. For example, a customer asks for a lipstick recommendation in Figure \ref{fig:illustration}. 

\vspace{-0.5em}
\paragraph{How conversation ends.}
When all remaining products from $P_o$ satisfy $preference$, the seller agent randomly recommends one of the products to the customer. The conversation continues for up to three more turns to conclude the shopping journey. Note that we do not specifically define finding the exact target product as a stopping criteria. Instead, we consider search results converging to a ``product family'' that satisfies $preference$ is close enough to satisfy customer's shopping interest.

\vspace{-0.5em}
\paragraph{Improving naturalness by adding hints.}
\label{sec:improving_naturalness}
Customers may not always be aware of available options for a given product aspect, which was not well-addressed in prior studies~\cite{zhang2018towards,bi2019conversational,zou2022learning}. 
It is more user-friendly to provide customers with options when asking clarification questions. For example, if the aspect under consideration is the size of a product's solid-state drive (SSD), a general customer may not know the typical storage size of SSDs. 
To mitigate this, we propose to always require the seller to provide hints based on up to 3 most frequent values associated with the product aspect being discussed (e.g., 256GB, 512GB, or 1TB for SSD size). 
Similar to the proposal of product aspects by the decision tree, we also dynamically retrieve the most frequent values of each product aspect from the remaining candidate set of products. Providing typical aspect values as examples can help clarify the meaning of the product aspect, especially when it may not be known to customers, without requiring two additional turns of clarification questions.

\vspace{-0.5em}
\paragraph{Improving naturalness by LLM knowledge.}
The aspects may not always be intuitive to the customer, since customers have different levels of knowledge and may not understand certain technical or specific product aspects. LLMs are known to have world knowledge internally, which can be used to ``smooth'' out conversations given non-intuitive product features. We encourage LLM-powered agents to clarify the meaning of aspects that require additional knowledge to be understood. We also allow the seller agent to reorder the decision tree-selected product features to be discussed, if multiple features are provided at the same time (See bolded prompts in Table \ref{tab:simple_prompt}, Appendix Tables  \ref{tab:seller_prompt}, \ref{tab:customer_prompt} \& \ref{tab:single_pass_prompt}). Note that the decision tree is used to globally select the best-k product aspects among all possible aspect values to be elicited from the customer. The LLM interprets and locally reorders selected product aspects to enhance the natural flow of the conversation.

\section{Experiments and Results}
\subsection{Experimental Setup}
\label{sec:setup}

We use a subset of the TREC product search dataset\footnote{\url{trec-product-search.github.io}}, which is a cleaned derivative of the Amazon product dataset \cite{mcauley2015image} as our product catalog. As we explained previously, we do not have any available real-user behavior data. We experiment with the Home \& Kitchen, Electronics, and Beauty \& Personal domains, which contain 135k, 54k, and 47k products respectively. 
We remove non-intuitive product aspects, such as ``ASIN'', ``Date First Available'' and ``Is Discontinued By Manufacturer'' and normalize certain aspects, such as ``Brand Name'' to ``Brand'' and ``Colour'' to ``Color''. 
We also normalize the price range and customer review scores into ranges in the format of ``between \$10 and \$20'' and ``higher than 4.5 stars'' respectively. Finally, we discard products with fewer than two features, excluding customer reviews and price ranges.


\subsection{Human Evaluation} 
\label{sec:human_evaluation}
Since there is no gold conversation to evaluate against as we explained \S\ref{sec:related_work}, we perform human evaluation at the conversation and utterance levels to examine the quality of the dialogues.
The metrics are summarized in Table \ref{tab:human_evaluation_metrics}. For conversation-level evaluations, judges are asked to score each conversation on a 5-point Likert scale. The definition of each score is shown in Appendix Table \ref{tab:score_interpretation}. Note that the \textit{overall naturalness} of 5 means the generated conversation is \textit{indistinguishable from real-life human conversations}. 
We also evaluate the conversations on a finer-grained utterance level where judges identify all unsatisfactory, non-script following utterances, as a binary labeling task. 

We evaluate conversations generated under 4 settings: GPT-4 single-pass generation, GPT-4 interactive generation, LLaMA-2-70B-Chat single-pass, and LLaMA-2-70B-Chat interactive generation. We begin by assessing inter-annotator agreement using Kendall's $\tau$. To achieve this, we provide detailed annotation instructions to expert judges divided into 5 groups, with 3 judges per group. Each judge is assigned 12 examples, spread across 3 domains with 4 configurations per example. Therefore, we have evaluation results of 180
instances for 60
conversations, allowing us to calculate inter-annotator agreement. Furthermore, we assign two of the judges to work on 24 additional examples each. These examples are divided into 3 domains, with 2 examples per domain, and each example involves 4 configurations. Thus, we evaluate 
108 dialogues in total. 

\begin{table}[t]
\small
\begin{center}
\setlength{\tabcolsep}{3pt} 
    \begin{tabular}{lp{0.6\linewidth} }
    \hline
\textbf{Metric} & \textbf{Description}\\ \hline
\textit{Conversation-Level} &  \\ \hline
Realism & Whether the conversation is likely to happen in the real world from a logical perspective. \\ \hline
Conciseness & Whether the conversation is too verbose. \\ \hline
Coherence & Whether the conversation is fluent and coherent with respect to the conversation history from the beginning. \\ \hline
Overall naturalness & Overall subjective impression of the conversation and whether it sounds natural. \\ \hline \hline
\textit{Utterance-level} &  \\ \hline
Realism & Whether the utterance is likely to be spoken in the real world from a logical perspective. \\ \hline
Script-following & Whether the utterances are consistent with the product features and customer preferences. \\ \hline
\end{tabular}
\vspace{-1em}
\caption{Human evaluation metrics.}
\label{tab:human_evaluation_metrics}
\end{center}
\vspace{-3em}
\end{table}


\subsection{Results} \label{sec:human_evaluation_results}
Table \ref{tab:conversation_level_human_evaluation} shows the average scores of conversational-level human evaluation results, and Table \ref{tab:utterance_level_human_evaluation} shows the average number of unsatisfactory utterances marked by the judges. 
As Table \ref{tab:conversation_level_human_evaluation} and \ref{tab:utterance_level_human_evaluation} show, we obtain a clear trend of scores across almost all metrics with high agreements (Average Kendall's $\tau=0.51$ \& $0.44$ respectively) 
for both conversation-level and utterance-level evaluations. Unsurprisingly, GPT-4 outperforms LLaMA-2-70B-Chat, which is consistent with the existing leaderboard\footnote{\url{https://huggingface.co/spaces/lmsys/chatbot-arena-leaderboard}}.
 
The single pass generation approach consistently outperforms the interactive generation for both LLM types. This is because the single pass approach takes the global generation plan for both seller and customer agents and performs dialogue generation with a single pass, thus there are less likely to be contextual conflicts. 
On the contrary, as our error analysis (\ref{sec:error_analysis}, coherence) \& Appendix Table \ref{tab:bad_coherence} show, in the interactive approach, the LLM-powered agents have no access to the generation plan beyond the next utterance. Due to the alternative generation paradigm, there is a greater potential for errors. The only exception is for LLaMA-2-70B-Chat, where the single pass approach has a higher average number of non-script-following utterances than the interactive setting. This is likely because LLaMA-2-70B-Chat tends to follow instructions less closely than GPT-4 when crafting a coherent conversation. 

\begin{table}[t] \small
\centering
\setlength{\tabcolsep}{2pt} 
\begin{tabular}{llllll}
\hline
\textbf{Model} & \textbf{Strategy} & \textbf{Real} & \textbf{Concise} & \textbf{Coherence} & \textbf{Natural}\\ \hline
GPT-4   & Single pass & \textbf{4.3} & \textbf{4.8} & \textbf{4.7} & \textbf{4.2}\\ \hline
GPT-4   & Interactive & 3.3 & 2.9 & 3.4 & 2.9 \\ \hline
LLaMA-2 & Single pass & 4.1 & 4.7 & 4.1 & 3.9\\ \hline
LLaMA-2 & Interactive & 2.5 & 3.0 & 3.3 & 2.5 \\ \hline

\end{tabular}
\vspace{-1em}
\caption{Conversational-level human evaluation results measured by Realism, Conciseness, Coherence and Naturalness. Average Kendall's $\tau = 0.51$.}
\label{tab:conversation_level_human_evaluation}
\vspace{-0.5em}
\end{table}

\begin{table}[t] \small
\centering
\setlength{\tabcolsep}{3pt} 
\begin{tabular}{llcc}
\hline
\textbf{Model} & \textbf{Strategy} & \textbf{Unrealistic $\downarrow$} & \textbf{Not Following $\downarrow$}\\ \hline
GPT-4 & single pass & \textbf{0.8} & \textbf{0.1}\\ \hline
GPT-4 & interactive & 2.0 & 0.4 \\ \hline
LLaMA-2 & single pass & 1.2 & 1.7 \\ \hline
LLaMA-2 & interactive & 2.9 & 1.1 \\ \hline

\end{tabular}
\vspace{-1em}
\caption{Utterance-level human evaluation results in terms of the number of labeled unrealistic and non-script-following utterances. The lower, the better. Average Kendall's $\tau = 0.44$.}
\label{tab:utterance_level_human_evaluation}
\vspace{-2em}
\end{table}

\begin{table}[t] \small
\centering
\setlength{\tabcolsep}{3pt} 
\begin{tabular}{cc}
\hline
\textbf{Total \# conv} & 3600\\ \hline
\textbf{Domains} & Electronics\\ 
 & Home \& Kitchen\\ 
 & Beauty \& Personal Care\\ \hline
\textbf{\# Conv per domain} & 1200\\ \hline
\textbf{avg \# utterance per conv} & 19.7\\ \hline
\textbf{avg \# search per conv} & 2.2\\ \hline
\textbf{avg generation time per conv} & 75.6s\\ \hline
\end{tabular}
\vspace{-1em}
\caption{Statistics of \textit{Wizard of Shopping} dataset.}
\label{tab:stats}
\vspace{-2em}
\end{table}

\subsection{Wizard-of-Shopping (WoS) Dataset}

Since human evaluation results validate that \method with GPT-4 single pass generation can generate high-quality conversations, we propose to release an official, larger scale \textit{WoS} dataset to (1) promote more research in CPS and (2) study the effectiveness of generated datasets in real downstream tasks. Table \ref{tab:stats} shows the statistics of our \textit{WoS} dataset.
A typical example of a generated conversation is shown in Appendix Table \ref{tab:good_example}. 
In contrast to prior conversational e-commerce datasets that are either large-scale but unnatural \cite{zou2022learning}, or natural but small-scale \cite{Bernard:2023:SIGIR}, our \textit{WoS} dataset consisting of 3600 conversations is large-scale, and is designed to be human-like and realistic. Further, since our \method approach can be easily applied to other product domains to generate additional conversations, we will continue to expand the \textit{WoS} dataset in terms of quantity and number of product domains\footnote{We will release the dataset and dialogue generation scripts upon acceptance.}.

\subsection{Error Analysis of \textit{WoS} Dataset} \label{sec:error_analysis}
 Despite the high naturalness and coherence of the conversations generated by TRACER, we identify the following errors in them. 
 
 \paragraph{Verbosity.} In the interactive generation setting, the seller agent often ``hurries'' to include all the assigned product aspects in a single lengthy utterance. This behavior is primarily due to the seller agent's inability to anticipate the potential for multiple turns in the conversation, such that it could gradually discuss one or two product aspects per utterance, as instructed in the prompt. Consequently, the interactive setting often produces much lengthier utterances compared to the single-pass generation, adversely impacting realism, conciseness, and naturalness scores. As the examples in Tables \ref{tab:bad_conciseness} \& \ref{tab:bad_coherence} show, the seller tends to produce overly lengthy utterances.

 \paragraph{Coherence.} We encourage the customer agent to spontaneously ask clarification questions in the conversation in order to improve the naturalness of the conversation. In interactive generation settings, when the customer asks clarification questions, the product value selection is usually left unfinished due to the customer's confusion. In such cases, the seller agent usually also successfully responds to the customer's questions. However, the seller does not give the customer a second chance to make a value selection response and instead directly moves to asking about new product aspects (See sentences in bold in Table \ref{tab:bad_coherence}). Consequently, this abrupt transition significantly undermines the coherence, realism, and naturalness of the conversation.

 \paragraph{Bad features.} The extracted aspect-value pairs from the product catalog are not perfect~\cite{putthividhya2011bootstrapped,raju2009unsupervised,yang2022mave}, with some of them being very unnatural, despite the manual cleaning we performed and explained in \S\ref{sec:setup}. When such unnatural or unrealistic product aspects are selected by the decision tree, the LLMs are forced to verbalize them. In such cases, certain utterances may look awkward. For example, the inappropriately selected product aspects (in bold) in Table \ref{tab:bad_feature} make the conversation sound awkward.

 \paragraph{Not following instructions.} The LLMs may not always follow the prompts to verbalize selected product features and preferences faithfully. For example, the seller agent may ask about product aspects that have not been explicitly assigned from the catalog, based on the commonsense understanding of the LLM. Similarly, a customer agent may invent requirements that are not present in the assigned product values from the catalog. Additionally, the customer agent may not be precise in conveying their preferences, sometimes responding with statements like ``this is an optional feature for me'' instead of ``I do not want this feature''. In Table \ref{tab:not_instruction_following}, all the bolded unassigned product features are invented by the LLM. 
 We quantify this issue in the ``Not Following'' column in Table \ref{tab:utterance_level_human_evaluation}.

\subsection{Comparison with Prior Dataset}\label{sec:mgshopdial_comparison}

This section compares \textit{WoS} to MG-ShopDial \cite{Bernard:2023:SIGIR} which is the only publicly available e-commerce dialogue dataset. 
\vspace{-0.5em}

\paragraph{Annotation Protocol.} MG-ShopDial is annotated by crowdsourcers, with some annotators acting as shopping assistants and others as customers. Each role is given a checklist of actions to complete, ensuring that the annotated dataset encompasses a variety of intents. While \textit{WoS} is initially generated by \method, we later demonstrate that through careful dialogue planning, our simulated dialogues can also encompass diverse intents. Additionally, because customer preferences are randomly sampled and vary in complexity, each dialogue may have a different ``checklist'', resulting in greater diversity at the corpus level compared to MG-ShopDial.
\vspace{-0.5em}
\paragraph{Product Catalog.} MG-ShopDial offers a curated catalog to annotators, which covers around 14 items per category. However, this limited selection may not accurately reflect the breadth of choices available in real-world shopping scenarios. As detailed in \S\ref{sec:setup}, we index over 236k product and the utterance generation in \textit{WoS} depends on the most recent search results. Therefore, the dialogue generation process of \method closely resembles the iterative ``search-and-refine'' procedure of typical shopping experiences.
\vspace{-0.5em}
\paragraph{Intent.} We use OpenAI GPT-4\footnote{gpt-4-0125-preview} to label each utterance with intents defined by \citet{Bernard:2023:SIGIR}. See more setup details in Appendix \ref{sec:intent_analysis}. As Table \ref{tab:intent} shows, similar to MG-ShopDial, our \textit{WoS} conversations exhibit diverse distributions of intents, which is consistent with the high naturalness observed in \S\ref{sec:human_evaluation_results}. Furthermore, MG-ShopDial has significantly more ``Recommend'' intent than \textit{WoS}, while \textit{WoS} has a dominating proportion of ``Elicit preferences'' intent. This correlates with our observation that the wizard (seller) in MG-ShopDial memorizes all candidate products due to a limited catalog size and starts the product recommendation stage early without explicit product search. In contrast, with the decision tree for dialogue planning, \textit{WoS} conversations aim to narrow down product aspect values for product search from a real product catalog and adopt the strategy of not selecting any product until the search is converged. As a result, \textit{WoS} conversations are significantly shorter than MG-ShopDial (19.7 vs. 34.3 utterances).


\vspace{-0.5em}
\paragraph{Conversation styles.} Since MG-ShopDial is generated by humans with keyboard typing, their sentences are usually more brief (8.5 words per utterance), with frequent absence of concluding punctuation. Specifically, only 256 of its 2,196 utterances conclude with a period, while 479 end with a question mark.  
In contrast, utterances in \textit{WoS} tend to be lengthier, averaging 22.8 words per utterance, characterized by near-perfect grammar (as illustrated in Appendix Table \ref{tab:good_example}). 
As a result, MG-ShopDial conversations tend to be more lengthy and casual than \textit{WoS}.

\paragraph{Expandability.} For integrating additional domains into MG-ShopDial, it is necessary to curate a new list of products and potentially retrain annotators. Our proposed \method can seamlessly extend to generating shopping dialogues for any new domains once the products are indexed.
\section{Downstream Tasks}
\label{sec:downstream_tasks}
In this section, we show the effectiveness of the Wizard-of-Shopping dataset by applying it to two downstream tasks: Conversational Query Generation (CQG) and Conversational Product Ranking (CPR). For both tasks, we divided the \textit{WoS} conversations into 3,000 for training, 300 for validation, and 300 for testing, with each split containing an equal number of dialogues from the three domains. More details are available in Appendix \ref{sec:detailed_downstream_tasks}.

\subsection{Conversational Query Generation (CQG)}

Real product search conversations are likely to be verbose and redundant for the purpose of product ranking.  
Similar to the conversational search~\cite{yu2020few,vakulenko2021question,chen-etal-2022-reinforced,wu2022conqrr} where a reformulation model is used to generate a more informative query from the dialogue history, we address \textit{conversational query generation} (CQG) for product search. 
CQG aims to extract essential information such as the product category under discussion, desired features, undesired features, and optional product preferences from the customer. This extracted information can then be used as a query for a product search engine. And in fact, this is a reverse task of the LLM verbalization where we extract user preferences from the shopping dialogues. Similar to Dialogue State Tracking~\cite{lu2023dialgen}, CQG is helpful in tracing the interest of customers during the conversation. 
As \citet{bi2019conversational} suggests, utilizing positive and negative product features are crucial for training a conversational product ranking model. We train a conversational query generator on the \textit{WoS} data to extract the product category and the customer-preferred product features:
\begin{equation}\label{eq:cqg}  \small
    [PC, Wanted, Unwanted, Optional] \gets CQG(Dialogue)
\end{equation}
\vspace{-2em}

\subsubsection{Approaches} \label{sec:conversational_query_generation_approaches}

\paragraph{Baseline.} As stated in \S\ref{sec:related_work}, to the best of our knowledge, \textit{WoS} is the only CPS dataset that can be used for training downstream tasks. Therefore as a simple baseline, we directly use the full conversation history as the predicted query.
\paragraph{Dialogue --> Query (D2Q).} We use a seq2seq model, Longformer Encoder-Decoder \citep[LED;][]{beltagy2020longformer} fine-tuned on our \textit{WoS} dataset to predict the product category and the \textit{wanted}, \textit{unwanted} and \textit{optional} product features, given the full conversation history.

\paragraph{D2Q (GPT-4).} For reference of LLM performance, we few-shot prompt OpenAI gpt-4-turbo-2024-04-09 with the same inputs and outputs as D2Q.


\begin{table}[t]  \small
\centering
\setlength{\tabcolsep}{4pt}
\begin{tabular}{llllll}
\hline
\textbf{Approach} & \textbf{F1} & \textbf{R-1} & \textbf{R-2} & \textbf{R-L}\\ \hline
Baseline & 0.008 & 0.137 & 0.047 & 0.087  \\  \hline  
D2Q & 0.834 & 0.899 & 0.822 & 0.873 \\ \hline
D2Q (GPT-4) & 0.553 & 0.793 & 0.628 & 0.734 \\ \hline
\end{tabular}
\vspace{-0.5em}
\caption{Conversational Query Generation performance at the final turn measured by exact-F1, ROUGE-1, -2, and -L.}
\label{tab:conversational_query_generation}
\vspace{-2em}
\end{table}

\subsubsection{Experimental Results}
Table \ref{tab:conversational_query_generation} shows the performance of the CQG task.
As expected, the weak baseline using conversation history performs poorly, while our fine-tuned query generator performs much better. 
Although GPT-4's performance is slightly underestimated because few-shot demonstration examples do not show every edge case, D2Q by \textit{WoS}-finetuned LED shows its superiority of both performance and cost in terms of both latency and price over GPT-4 as an example of LLM.

\subsection{Conversational Product Ranking (CPR)}
As the core component of CPS, CPR directly ranks the candidate products given the shopping conversation. We index or embed each product's title and concatenate their aspect-value pairs according to the following approaches:

\subsubsection{Approaches}

\paragraph{Baseline} 
We directly feed the full conversation history as the query to a BM25 ranker.

\vspace{-0.5em}
\paragraph{Dialogue --> Query --> Product (D2Q2P)} We use D2Q setting in \S\ref{sec:conversational_query_generation_approaches} to generate queries given the full conversation history and apply the queries to a ranker, i.e. \textit{Products = Ranker(CQG(Dialogue))}. We experiment with both a sparse ranker, BM25, and a dense ranker \citep[fine-tuned RoBERTa;][]{liu2019roberta}.

\paragraph{D2Q2P (GPT-4)} We use D2Q (GPT-4) setting in \S\ref{sec:conversational_query_generation_approaches} and feed the predicted query to a BM25 ranker.

\begin{table}[t]  \small
\centering
\setlength{\tabcolsep}{2pt} 
\renewcommand{\arraystretch}{1.0} 
\begin{tabular}{llllllll}
\hline
\textbf{Approach} & \textbf{Ranker} & \textbf{MRR} & \textbf{H@1} & \textbf{H@10} & \textbf{H@100}\\ \hline
Baseline & BM25& 0.162 & 0.107 & 0.203 & 0.587 \\ \hline \hline
D2Q2P & BM25& \textbf{0.838} & \textbf{0.767}  & \textbf{0.927} & \textbf{0.993} \\
D2Q2P & Roberta & 0.675 & 0.583  & 0.780 & 0.937 \\ \hline
D2Q2P (GPT-4) & BM25 & 0.763 & 0.680 & 0.903 & 0.903 \\ \hline 

\end{tabular}
\vspace{-0.5em}
\caption{Downstream CPR performance at the final turn.}
\label{tab:ranker}
\vspace{-2em}
\end{table}

\subsubsection{Experimental Results}
Table \ref{tab:ranker} shows the mean reciprocal rank (MRR) and Hit$@k$ of all methods. 
Similar to the trend observed for CQG, our D2Q2P approach significantly outperforms the baseline and GPT-4. Interestingly, the sparse BM25 ranker greatly outperforms the dense RoBERTa-based ranker. This is because the product representations include feature names that are lexically similar to the gold queries. Consequently, BM25 exhibits a strong performance in this ranking task. Moreover, the dense ranker may be under-fitted. The query generators and dense rankers fine-tuned on our \textit{WoS} dataset significantly outperform the baseline that is not trained on our dataset.

\section{Conclusions}
We propose a method, \method, to automatically generate target-oriented shopping conversations without any human annotations. 
We leverage decision tree to explore the vast product search space, and construct a dialogue plan that minimizes the number of search steps required to retrieve a relevant product.
The resulting corpus (\textit{WoS}), generated using single-pass approach with GPT-4, not only achieved highly natural (4.2/5.0) and coherent (4.7/5.0) ratings from human annotators, but also showed substantial improvements when applied to two downstream tasks. By releasing our dataset and approach, we hope to expedite the research and development of intelligent CPS systems in future.


\section*{Limitations}
Despite the demonstrated potential of our \method approach and the \textit{WoS} dataset, there are still some limitations that we leave for future work. 
First, \method does not consider more complicated shopping behavior such as comparison among similar products~\cite{vedula2022matters}. 
In the future, we could extend the ability of \method and allow customers to choose a similar product in the search results for comparison with assistance from a LLM agent. 
Second, the quality of \textit{WoS} conversations relies on the quality of the product catalog. Since noise in the product catalog can directly propagate to the generated conversations. Methods proposed to curate and clean the product catalog~\cite{ghani2006text,yang2022mave,vedula2022matters} could be applied to further clean a product catalog. 
Finally, since the decision tree treats every product aspect and value as a categorical label, the semantics of product features are under-utilized. Future work may further leverage the zero-shot ability of LLMs to mine the semantics of the product aspects, so that similar product features may be merged to produce a more natural and efficient dialogue plan.

\section*{Ethical Statements}
The potential risk of our work could be biased simulations of real-world users' behavior, such that the learned downstream CQG and CPR models are biased. As a result, the end user's will of shopping might be distorted by the CPS system. However, we note that this risk is hypothetical.
\clearpage
\bibliography{sample-base}
\bibliographystyle{acl_natbib}
\clearpage
\appendix

\begin{table*}[t]
\small
\centering
\begin{tabular}{lll}
\hline
\textbf{Metric} & \textbf{Score} & \textbf{Interpretation} \\ \hline
Realism & 1 & Impossible to happen in the real world. \\
& 2 & Unlikely to happen in the real world, but still possible. \\
& 3 & May happen in the real world, but unrealistic. \\
& 4 & Likely to happen in the real world, with minor issue. \\
& 5 & Very likely to happen in real life. \\ \hline
Conciseness & 1 & Utterances are way too long and verbose. Not acceptable. \\
& 2 & Utterances are very long and verbose. \\
& 3 & Utterances are long but still acceptable. \\
& 4 & There is a space for simplifying utterances. \\
& 5 & Utterances cannot be further simplified. \\ \hline
Coherence \& Fluency & 1 & The conversation is not complete till the end. \\
& 2 & Contains serious coherence or fluency problems, but the conversation is complete.\\
& 3 & Contains obvious coherence or fluency problems. \\
& 4 & Contains minor coherence or fluency problems. \\
& 5 & Coherent and fluent conversation without any flaw. \\ \hline
Overall naturalness & 1 & Very unnatural conversation. Difficult to read. \\
& 2 & Contains obvious naturalness issues. \\
& 3 & Obviously this is machine-generated, with minor naturalness issues. \\
& 4 & No obvious issues found, but it's not as natural as humans. \\
& 5 & Good conversation. Indistinguishable from real-life conversations. \\
\hline

\end{tabular}

\caption{Interpretation of conversational-level human evaluation scores.}
\label{tab:score_interpretation}
\end{table*}


\section{More Related Work}
\subsection{Dialogue Generation with LLMs}

Collecting dialogue datasets for conversational applications requires significant effort from crowd workers~\cite{ 
budzianowski-etal-2018-multiwoz, wu-etal-2020-tod}, incurring substantial financial and time costs. Hence, alternative simulation approaches such as DIALOGIC were developed for dialogue generation using GPT-3 \cite{brown2020language} on the MultiWOZ dataset \cite{han2021multiwoz}.
It successfully generated dialogues with near-human fluency, even when trained with as few as 85 initial seed dialogues.
Similarly, \citet{li-etal-2023-autoconv} proposed to fine-tune OPT-13B~\cite{zhang2022opt} to generate conversations grounded on a given document. 
A semi-automated dialogue generation framework based on ChatGPT~\cite{lu2023dialgen} was proposed to generate complex conversations in the insurance domain by generating sub-dialogues and using human feedback to ensure consistency.
They demonstrated that incorporating the generated dataset into existing auto insurance claim call datasets can enhance the performance of Dialogue State Tracking (DST) \cite{young2010hidden}. 


Unlike previous methods, we do not rely on initial seed annotations; we only leverage the zero-shot capabilities of LLMs to generate conversations that are grounded in a product catalog and a simulated user profile. 
Once a user profile is synthesized from the product information and elicited customer preferences, a decision tree is fitted to guide the dialogue generation, ensuring that the conversation concludes with the discovery of the target product. 
Similar to the DST task~\cite{lu2023dialgen}, we demonstrate that our generated dataset can be used to train a query generator that extracts preferences for product attribute values (\S\ref{sec:downstream_tasks}).

\section{Detailed Human Evaluation Metric Descriptions}
\subsection{Conversation-level Evaluations} 
The metrics are summarized in Table \ref{tab:human_evaluation_metrics}. For conversation-level evaluations, judges are asked to score each conversation on a Likert scale of 1 to 5. The meaning of each of the scores is interpreted in Table \ref{tab:score_interpretation}. 

\paragraph{Realism.} To distinguish counter-intuitive conversations, \textit{realism} examines whether the conversation is likely to happen in the real world from a logical perspective. We also consider \textit{Factuality}, which captures the commonsense-level factual errors, since they essentially reflect similar criteria. Meanwhile, we consider whether the speakers stick to their own role, i.e. the seller should ask questions or recommend products but not buy products, while the customer should focus on answering proposed questions to narrow down products to buy, instead of recommending products. 
We combine these three aspects into one score since they are highly related.

\paragraph{Conciseness} examines whether the conversation is too verbose, such as a single utterance contains too many aspects to discuss.

\paragraph{Coherence} measures whether the conversation is coherent to the conversation history from the beginning. Additionally, we also consider fluency, as it is also closely relevant. Specifically, we examine (1) whether there is repetition or forgetting about the context; (2) whether the agents ignore the previous utterance and proceed to the next topic; (3) whether the conversation gets stuck at some point and never proceeds.

\paragraph{Overall naturalness} reflects the overall subjective impression of the conversation. Opposing to the \textit{realism} which reflects whether the conversation is logical, the naturalness asks whether the judge subjectively feels humanness from the conversation.

\subsection{Utterance-level Evaluations}
We further evaluate the conversations on a finer-grained scale. Judges are asked to list all unsatisfactory utterance IDs, as a binary classification task.

\paragraph{Realism.} Similar to the conversation level, we examine the realism, factuality, and consistency of the role at the utterance level.

\paragraph{Script-following.} Given the assigned product aspects and values, as well as the customer's preference (\textit{wanted}, \textit{optional}, or \textit{unwanted}), we evaluate whether the utterances are consistent with the assigned features. The utterances should cover all assigned features and should not alter the type of the customer's preference (e.g. switching \textit{unwanted} to \textit{optional}). Further, agents should not invent new product features that are not assigned, which may create inconsistency with the search plan and the semantics of the conversation.

\paragraph{Reasonable features.} The selected product features by the decision tree may contain inappropriate or unnatural product aspects and values to be discussed. For example, the customer is not supposed to know the value of the aspect ``model number''. This issue usually originates from the quality of the product metadata, which is impossible to perfectly curate. Table \ref{tab:utterance_level_human_evaluation} shows the results excluding utterances with unreasonable features to avoid confusion.


\section{Large Language Models for Dialogue Generation} \label{sec:llm}
We experiment with two different LLMs: OpenAI GPT-4\footnote{\url{https://platform.openai.com/docs/models}, gpt-4-0613} and LLaMA-2-70B-chat. As the \textit{chatbot-arena-leaderboard}\footnote{\url{https://huggingface.co/spaces/lmsys/chatbot-arena-leaderboard}} shows, GPT-4 was the best-performing LLM by then.
However, as GPT-4 is proprietary, we also experiment with LLaMA-2-70B-chat, which is one of the best performing commercially available instruction-tuned LLMs. Due to the high GPU memory requirement of LLaMA-2, we use a 5-bit quantized version.\footnote{\url{https://github.com/abetlen/llama-cpp-python}, \url{https://huggingface.co/TheBloke/Llama-2-70B-Chat-GGML/resolve/main/llama-2-70b-chat.ggmlv3.q5_K_M.bin}}

\begin{table*}[t]  \small
\centering
\renewcommand{\arraystretch}{1.0} 
\begin{tabular}{ll}
\hline
\textbf{Intent} & \textbf{Definition}\\ \hline
Greetings & Indicates the beginning or end of the conversation \\ \hline
Interaction structuring & Utterances that make the conversation structured and natural (e.g., thanking, stalling) \\ \hline
Disclose & The client discloses information about what they are looking for  \\ \hline
Clarification question & The agent asks a question to make sure it understands correctly a previous statement \\ \hline
Other question & Asks a question that is not a clarification question (e.g., factoid, follow-up questions) \\ \hline
Elicit preferences & The agent asks a question to find the client’s preferences (e.g., the color of an item, the budget) \\ \hline
Recommend & The agent recommends one or several items to the client \\ \hline
Answer & A participant gives an answer to the other participant’s information request \\ \hline
Explain & Provides an explanation to a previous statement (e.g., justifies suggestion or rejection of an item) \\ \hline
Positive feedback & Expresses positive feedback (e.g., confirmation, accept a recommendation)  \\ \hline
Negative feedback & Expresses negative feedback (e.g., disagreement, rejection of a recommendation) \\ \hline
Other & Does not fit other labels \\ \hline
\end{tabular}
\caption{Utterance intents defined by MG-ShopDial \cite{Bernard:2023:SIGIR}.}
\label{tab:intent_definitions}
\end{table*}

\begin{table*}[t]
\small
\begin{center}
    \begin{tabular}{p{0.95\linewidth} }
    \hline
    \textbf{Instruction} \\ 
    Given an utterance with context in a conversation under e-commerce shopping scenario, 
identify all intents of the utterance defined below (format of "intent: explanation"): \\ \hdashline

Greetings: Indicates the beginning or end of the conversation \\
Interaction structuring: Utterances that make the conversation structured and natural (e.g., thanking, stalling)\\
Disclose: The client discloses information about what they are looking for\\
Clarification question: The agent asks a question to make sure it understands correctly a previous statement\\
Other question: Asks a question that is not a clarification question (e.g., factoid, follow-up questions)\\
Elicit preferences: The agent asks a question to find the client’s preferences (e.g., the color of an item, the budget)\\
Recommend: The agent recommends one or several items to the client\\
Answer: A participant gives an answer to the other participant’s information request\\
Explain: Provides an explanation to a previous statement (e.g., justifies suggestion or rejection of an item)\\
Positive feedback: Expresses positive feedback (e.g., confirmation, accept a recommendation)\\
Negative feedback: Expresses negative feedback (e.g., disagreement, rejection of a recommendation)\\
Other: Does not fit other labels\\ \hdashline

Examples (Utterance followed by intents): \\
\{\} \\

Contexts: \{\} \\

Utterance: ``\{\}'' \\
Answer all plausible intents names of the utterance above, separate with ``;''. Do not return explanations.\\
Answer: \\ \hline
    \end{tabular}
    \caption{Prompt format for predicting utterance intents. } \label{tab:intent_prompt}
\end{center}
\end{table*}

\begin{table*}[t]  \small
\centering
\renewcommand{\arraystretch}{1.0} 
\begin{tabular}{lccc}
\hline
\textbf{Intent\%} & \textbf{Annotated MG-ShopDial} & \textbf{Predicted MG-ShopDial} & \textbf{Wizard of Shopping}\\ \hline
Answer & 17.0 & 3.4 & 0.1\\ \hline 
Clarification question & 11.8 & 9.3 & 3.0\\ \hline 
Explain & 11.5 & 6.7 & 5.5 \\ \hline 
Interaction structuring & 11.3 & 12.1 & 15.4\\ \hline 
Other question & 10.6 & 10.0 & 7.7 \\ \hline 
Positive feedback & 8.3 & 12.1 & 7.1 \\ \hline 
Greetings & 7.2 & 5.9 & 3.8 \\ \hline 
Recommend & 7.2 & \textbf{11.9} & \textbf{4.2} \\ \hline 
Disclose  & 6.4 & 15.3 & 15.6 \\ \hline 
Elicit preferences & 5.7 & \textbf{9.3} & \textbf{34.2}\\ \hline 
Other & 1.7 & 0.7 & 0\\ \hline 
Negative feedback & 1.4 & 3.3 & 3.4\\ \hline 

\end{tabular}
\caption{Intent distribution of utterances in MG-ShopDial \cite{Bernard:2023:SIGIR} and Wizard of Shopping.}
\label{tab:intent}
\end{table*}

\section{Intent Analysis of Wizard of Shopping vs. MG-ShopDial}
\label{sec:intent_analysis}
\paragraph{Experimental settings.} Extending \ref{sec:mgshopdial_comparison}, to compare the differences between utterances in \textit{WoS} and MG-ShopDial \cite{Bernard:2023:SIGIR}, we compare the intent distribution between these two datasets by prompting OpenAI GPT-4\footnote{gpt-4-turbo-2024-04-09} (see prompts in Table \ref{tab:intent_prompt}) to label each utterance with intents defined by \citet{Bernard:2023:SIGIR} (Table \ref{tab:intent_definitions}). Similar to \citet{Bernard:2023:SIGIR}, each utterance may correspond to multiple intent labels. We sample 100 \textit{WoS} conversations from each of the three domains for analysis. As \citet{Bernard:2023:SIGIR} indicated, their agreements among crowd workers are relatively low (Fleiss $\kappa$=0.187), thus we predict MG-ShopDial utterances with GPT-4 as well to make a fair comparison.


\section{Detailed Downstream Experiments}
\label{sec:detailed_downstream_tasks}
In this section, we extend \S\ref{sec:downstream_tasks} by comparing conversational-level and utterance-level generation approaches in addition to the weak baseline.

\subsection{Conversational Query Generation} \label{sec:detailed_conversational_query_generation}
\subsubsection{Approaches}
\paragraph{Baseline.} As we explained in \S\ref{sec:related_work}, to the best of our knowledge, our \textit{WoS} dataset is the only CPS dataset that can be used for training downstream tasks. Therefore as a simple baseline, we directly use the full conversation history as the predicted query.
\paragraph{Conversational-level.} We use a seq2seq model fine-tuned on our \textit{WoS} dataset to predict the product category and the \textit{wanted}, \textit{unwanted} and \textit{optional} product features, given the full conversation history.
\paragraph{Utterance-level.} We assume that each utterance between the seller and customer encodes one or more product features. We use a seq2seq model to extract product features from each utterance and concatenate the features to be the final query.
\paragraph{GPT-4} We few-shot prompt OpenAI gpt-4-0125-preview with the same inputs and outputs as the conversational-level setting above.

\subsubsection{Experimental Settings}
We fine-tune BART-base \cite{lewis-etal-2020-bart} for the utterance-level approach and use LED-base \cite{beltagy2020longformer} for the conversation-level approach due to the longer context required as input. 
We use a default learning rate of 5e-5 and train for 5 epochs for both models. 

\subsubsection{Experimental Results}
Table \ref{tab:detailed_conversational_query_generation} shows the performance of the CQG task. We report results on two settings: \textit{wanted} feature-only setups concatenate PC and wanted features and all feature setups concatenate all features as defined in Eq.~\ref{eq:cqg}. As expected, the weak baseline using the conversation history performs poorly, while our trained query generator performs much better.  
When comparing the utterance-level approaches, LED outperforms BART, since the training data for BART is slightly smaller than LED's. This is because we discard the input augmented conversation history that is too long for BART's maximum context window (1024 tokens). Also, the LED-based utterance level approach under-performs the conversation level approach, presumably because integrating the results from multiple inferences passes is more error-prone. 

We also experiment with GPT-4 as CQG for reference. Since the few-shot demonstration examples do not show every edge case, GPT-4's performance is slightly underestimated. Nonetheless, D2Q by \textit{WoS}-finetuned LED shows the superiority of both performance and cost over GPT-4.
\subsection{Conversational Product Ranking}

\subsubsection{Approaches}
\paragraph{Baseline} Similar to the CQG task, we assume there is no conversational data available for training a ranker. Therefore, we directly feed the full conversation history as the query to a BM25 ranker.

\paragraph{Dialogue --> Product (D2P)} We directly embed the full conversation history to a dense ranker to rank the candidate products, i.e. \textit{Products = Ranker(Dialogue)}. 

\paragraph{Dialogue --> Query --> Product (D2Q2P)} We leverage D2Q approach to generate queries given the full conversation history and apply the queries to a ranker, i.e. \textit{Products = Ranker(CQG(Dialogue))}. We experiment with both a sparse ranker, BM25, and a dense ranker.

\paragraph{D2Q2P (GPT-4)} We use GPT-4 setting in \S\ref{sec:conversational_query_generation_approaches} and feed the predicted query to a BM25 ranker.

\subsubsection{Experimental Settings}
For the dense rankers, we fine-tune a Longformer-base \cite{beltagy2020longformer} for the D2P setting, and a RoBERTa-based ranker\footnote{\url{https://huggingface.co/castorini/ance-msmarco-passage}} for the D2Q2P setting using a dense passage retrieval \cite{karpukhin-etal-2020-dense} training approach \footnote{\url{https://haystack.deepset.ai/tutorials/09_dpr_training}}. For training both D2P and D2Q2P dense rankers, we use the ranking results given the corresponding gold queries as positive examples. The hard negative examples are constructed by removing the positive examples from the search results of the target product category.

\subsubsection{Experimental Results}
Table \ref{tab:detailed_ranker} shows the mean reciprocal rank (MRR) and Hit$@k$ of all methods. 
We observe that the D2P approach only slightly outperforms the baseline approach without training. We suspect that our DPR-Longformer is still under-fitted given 3000 conversations as the training set. On the contrary, by leveraging the query generators trained on our collected conversations, the ranking performance is greatly improved. 
Similar to the trend observed in the CQG task, the conversation-level approach outperforms the utterance-level approach that integrates multiple passes of inference outputs.

\begin{table}[t]  \small
\centering
\setlength{\tabcolsep}{3pt}
\begin{tabular}{lllllll}
\hline
\textbf{Approach} & \textbf{QG} & \textbf{Feat.} & \textbf{F1} & \textbf{R-1} & \textbf{R-2} & \textbf{R-L}\\ \hline
Baseline & - & + & 0 & 0.056 & 0.020 & 0.048 \\ 
Baseline & - & +/-/? & 0.008 & 0.137 & 0.047 & 0.087  \\  \hline \hline
Utterance & BART & + & 0.698 & 0.592 & 0.482 & 0.562 \\ 
Utterance & BART & +/-/? & 0.656 & 0.664 & 0.489 & 0.592 \\  \hline
Utterance & LED & + & 0.704 & 0.610 & 0.509 & 0.578 \\ 
Utterance & LED & +/-/? & 0.748 & 0.720 & 0.561 & 0.657 \\  \hline
Conversation & LED & + & 0.834 & 0.899 & 0.822 & 0.873 \\ 
Conversation & LED & +/-/? & 0.873 & 0.900 & 0.768 & 0.840 \\  \hline \hline
Conversation & GPT-4 & + & 0.553 & 0.793 & 0.628 & 0.734 \\  
Conversation & GPT-4 & +/-/? & 0.669 & 0.833 & 0.584 & 0.703 \\ \hline
\end{tabular}

\caption{Detailed Conversational Query Generation (QG) performance at the final turn measured by exact-F1, ROUGE-1, ROUGE-2, and ROUGE-L. We report the performance of only using the desired features (+) for downstream ranking, as well as using all features (wanted (+), unwanted (-), and optional (?)). Detailed version of Table \ref{tab:conversational_query_generation}.}
\label{tab:detailed_conversational_query_generation}
\end{table}

\begin{table}[t]  \small
\centering
\setlength{\tabcolsep}{0.8pt} 
\renewcommand{\arraystretch}{1.0} 
\begin{tabular}{llllllll}
\hline
\textbf{Approach} & \textbf{QG} & \textbf{Ranker} & \textbf{MRR} & \textbf{H@1} & \textbf{H@10} & \textbf{H@100}\\ \hline
Baseline & - & BM25& 0.162 & 0.107 & 0.203 & 0.587 \\ \hline \hline
D2P & - & longformer & 0.201 & 0.143  & 0.247 & 0.620 \\ \hline
Utterance & BART & BM25& 0.629 & 0.540  & 0.733 & 0.900 \\
Utterance & BART & Roberta& 0.217 & 0.163 & 0.270 & 0.477\\ \hline
Utterance & LED & BM25& 0.667 & 0.593 & 0.740 & 0.900 \\
Utterance & LED & Roberta& 0.207 & 0.163 & 0.250 & 0.460 \\ \hline
Conversation & LED & BM25& \textbf{0.838} & \textbf{0.767}  & \textbf{0.927} & \textbf{0.993} \\
Conversation & LED & Roberta & 0.675 & 0.583  & 0.780 & 0.937 \\ \hline \hline
Conversation & GPT-4 & BM25 & 0.763 & 0.680 & 0.903 & 0.903 \\ \hline 
\end{tabular}
\caption{Detailed downstream Conversational Product Ranking performance at the final turn. QG: query generator. Detailed version of Table \ref{tab:ranker}.}
\label{tab:detailed_ranker}
\end{table}


\begin{table*}[t]
\small
\begin{center}
    \begin{tabular}{p{0.95\linewidth} }
    \hline
    \textbf{System Prompt} \\ 
    You are acting as an Amazon seller to help narrow down product requirements for the customer that you are chatting with. Your output utterance is to clarify the customer's precise need by asking questions about the requested product features.  \\ \hdashline
    \textbf{Conversational history} \\ 
    Conversational history:  \\ 
    seller: \{\{\textit{Utterance\_S1}\}\} \\ 
    customer: \{\{\textit{Utterance\_C1}\}\} \\ 
    ...... \\
    \hdashline
    \textbf{Instructions about Features} \\ 
    First of all, \textbf{if you are asked clarification questions by the customer, you must friendly address the customer's questions first.} In this case, you must give the customer an opportunity to answer unanswered questions before you proceed. This means you may disregard the following new features. Otherwise, you must ask the customer about the following specific aspect of the requested product category provided by the output of the ``update\_dialogue\_state'' function. In your response, pick one or two aspects to ask from ``remaining\_feature'', so that the conversation is most natural, e.g. asking about general aspects first and then diving into detailed aspects. Once these one or two aspects are picked, your questions must strictly cover all of the words in the aspects or aspect value pairs. \textbf{Integrate the examples and values provided naturally into your utterance.} Do not ask about other aspects not covered above. If there is a conflict between your commonsense and the following aspect value pairs, you must prioritize the following provided aspect. You must ask in a concise and natural way. Do not repeat what you, the seller, have already said before. Keep the question brief. Do not talk about the budget unless you are asked to. Do not repeat the product name/category over and over.\\
    \hdashline
     \textbf{Instructions} \\ 
    What is the next utterance for you as a seller? Only reply with the response that you are supposed to make, not the customer's response. Don't start with your role, i.e. seller, just reply with the response. Remember, as a seller, you want to ask clarification questions to the customer. You never want to get products recommended or buy products from the customer. Be friendly to the customer, so that if the customer has any questions about the product or product aspect, answer patiently. \textbf{Treat the customer as if an old grandma has a limited understanding of the products.} \\
    \hdashline
    \textbf{Function Call Templates} \\
    Function name: update\_dialogue\_state \\
    Parameters: \{ \\
    \hspace{1cm} \textbf{remaining\_features: \{\{$Aspect_1: Expl_1, Expl_2, Expl_3$\}\},...} \\
    \hspace{1cm} mentioned\_positive\_features: \{\{\textit{$Aspect_2: Value_2$}\}\},... \\
    \hspace{1cm} mentioned\_optional\_features: \{\{\textit{$Aspect_3: Value_3$}\}\},... \\
    \hspace{1cm} mentioned\_negative\_features: \{\{\textit{$Aspect_4: Value_4$}\}\},... \\
    \}\\
\hline
    \end{tabular}
    \caption{Prompt format for interactive seller utterance generation. Prompts in bold are designed for improving conversation naturalness (\S\ref{sec:improving_naturalness}).} \label{tab:seller_prompt}
\end{center}
\end{table*}

\begin{table*}[t]
\small
\begin{center}
    \begin{tabular}{p{0.95\linewidth} }
    \hline
    \textbf{System Prompt} \\ 
    You are a \textbf{lazy} customer on Amazon chatting with a seller.  \\ \hdashline
    \textbf{Persona} \\ 
    \textbf{You have a limited understanding of the product to buy, so ask the seller about what the aspects mean when encountering a product aspect that an old grandma may not understand.} \\ \hdashline
    \textbf{Conversational history} \\ 
    Conversational history:  \\ 
    customer: \{\{\textit{Utterance\_C1}\}\} \\ 
    seller: \{\{\textit{Utterance\_S1}\}\} \\ 
    ...... \\
    \hdashline
    \textbf{Instructions} \\ 
    Given the utterance you just made, make an update to the specific \\ 
    aspects provided by the output of the update\_dialogue\_state \\
    function. For the aspect or aspect value pairs under ``remaining\_positive\_features'' below, you were asked to explicitly say that you like them in a \textbf{natural way}, so here you must pass these required aspect or aspect value pairs covered by your utterance under the ``mentioned\_positive\_features'' argument of function update\_dialogue\_state. Similarly, for the aspect or aspect value pairs under "remaining\_optional\_features" below, you were explicitly asked to express that they are optional or you are not concerned about them in a natural way, so here you must pass these optional aspect or aspect value pairs covered by your utterance under the ``mentioned\_optional\_features'' argument of function ``update\_dialogue\_state''. Finally, for the aspect or aspect value pairs under ``remaining\_negative\_features'' below, you were explicitly asked to express that you don't want them in a natural way, so here you must pass these negative aspect or aspect value pairs covered by your utterance under the ``mentioned\_negative\_features'' argument of functional update\_dialogue\_state. \\
    \hdashline
    
    \textbf{Function Call Templates} \\
    Function name: update\_dialogue\_state \\
    Parameters: \{ \\
    \hspace{1cm} remaining\_positive\_features: \{\{\textit{$Aspect_1: Value_1$}\}\},... \\
    \hspace{1cm} remaining\_optional\_features: \{\{\textit{$Aspect_2: Value_2$}\}\},... \\
    \hspace{1cm} remaining\_negative\_features: \{\{\textit{$Aspect_3: Value_3$}\}\},... \\
    \hspace{1cm} mentioned\_positive\_features: \{\{\textit{$Aspect_4: Value_4$}\}\},... \\
    \hspace{1cm} mentioned\_optional\_features: \{\{\textit{$Aspect_5: Value_5$}\}\},... \\
    \hspace{1cm} mentioned\_negative\_features: \{\{\textit{$Aspect_6: Value_6$}\}\},... \\
    \}\\ \hline
    \end{tabular}
    \caption{Prompt format for interactive customer utterance generation. Prompts in bold are designed for improving conversation naturalness (\S\ref{sec:improving_naturalness}). } \label{tab:customer_prompt}
\end{center}
\end{table*}

\begin{table*}[t]
\small
\begin{center}
    \begin{tabular}{p{0.95\linewidth} }
    \hline
    \textbf{Instruction} \\ 
    You are a scriptwriter. For the aspects or aspect value pairs below, write a chat conversation between the customer and seller about the seller trying to narrow down the customer's need. \textbf{The seller knows well about the sold products, while the customer has limited knowledge about the products.} This means the customer should not voluntarily propose what he wants, but passively respond to the aspect-values that the seller proposes. Therefore, the seller proactively asks clarification questions while the customer clarifies the needs. The customer should speak out first and say he/she wants to buy \{\{\textit{ProductCategory}\}\}. The seller asks about the aspect and the customer makes clarification on whether that aspect is something he/she wants, or it is optional. You MUST cover ALL aspect-value pairs below, including the wanted, unwanted, and optional aspect-value pairs. \textbf{Make sure you reorganize and reorder the aspect value pairs and translate them into a natural conversation in a meaningful way.} Start by talking about general product features and gradually transit into product-specific features. The seller should not ask too many aspects in one question. Instead only mention one or two aspects per question. Do not invent new aspects that are not mentioned below to talk about. Your script must be coherent with the conversation history, and there might be more scripts to be continued in the future, so don't terminate the conversation. Only output utterances, no non-verbal actions.  \\ \hdashline
    \textbf{Wanted / Optional / Unwanted Features} \\ 
  Aspect or aspect value pairs that the customer must say \{he wants to have / is optional / he does not want to have\} for the target product:  \\ 
  Aspect: \{\{\textit{Aspect}\}\}, Value: \{\{\textit{Value}\}\}; \\ 
  ...... \\\hdashline
  \textbf{Candidate Value Examples} \\ 
    \textbf{Additionally, there are some common values for each aspect to be mentioned. The seller should use these typical values in his question:} \\
    Aspect: \{\{\textit{Aspect}\}\}, Typical Values: \{\{\textit{$Value_1$}\}\}, \{\{\textit{$Value_2$}\}\}, \{\{\textit{$Value_3$}\}\}; \\ 
  ...... \\\hdashline
  \textbf{Format Instruction} \\
  Output in the format of \\
customer: AAAAA \\
seller: BBBBB \\
customer: CCCCC \\
seller: DDDDD \\ \hdashline
\textbf{Final Instruction} \\
Output a conversation following all instructions above and make sure you double check if ALL aspect value pairs above are mentioned: \\ \hline
    \end{tabular}
    \caption{Prompt format for single pass dialogue generation approach. Prompts in bold are designed for improving conversation naturalness (\S\ref{sec:improving_naturalness}).} \label{tab:single_pass_prompt}
\end{center}
\end{table*}

\begin{table*}[t]
\small
\begin{center}
    \begin{tabular}{l p{0.72\linewidth} }
    \hline
    \textbf{Product Category} & lipstick \\ \hline
    \textbf{Wanted Features} & color: dynamite red \\ \hline
    \textbf{Optional Features} & skin type; customer review; finish type; price; special feature \\ \hline
    \textbf{Unwanted Features} & brand: Gocheaper \\ \hline 
    \end{tabular}
    \begin{tabular}{l p{0.8\linewidth} }
    \hline
    \textbf{Speaker} & \textbf{Utterance} \\
    \hline
    customer& Hi there, I'm looking to buy a \textbf{lipstick} but I'm not quite sure where to start. \\\hdashline
    seller& Hello! Sure, I'd be happy to help you find the perfect lipstick. Let's start with the \textbf{color} you prefer. Popular ones are \textbf{red}, \textbf{black} or \textbf{clear}.\\\hdashline
    customer& Oh, I think I'd like \textbf{dynamite red}. \\\hdashline
    seller& Dynamite red sounds fabulous! Now let's talk about the \textbf{brand}. Some popular ones include \textbf{Revlon}, \textbf{NYX Professional Makeup}, and \textbf{Maybelline New York}. I noticed that there's also a brand named \textbf{Gocheaper}. Any preference regarding the brand?\\\hdashline
    customer& Oh, I \textbf{don't want} the brand \textbf{Gocheaper} for sure. As for the other brands, I have limited knowledge about them so I have no preference.\\\hdashline
    seller& That's perfectly fine. Now, on to the next aspect which is the \textbf{form} of the lipstick. It comes in \textbf{cream}, \textbf{stick} or \textbf{liquid} form. Do any of these appeal to you?\\\hdashline
    customer&  I'd prefer \textbf{not} to have a \textbf{liquid lipstick}. \\\hdashline
    seller& Noted, we'll avoid the liquid lipsticks. Now lipsticks come in different \textbf{finishes}. Some have a \textbf{glossy} finish, other have a \textbf{cream} finish and there are also \textbf{matte} ones. Do you have a preference for the finish type?\\\hdashline
    customer& \textbf{\textit{Oh, is that like how shiny the lipstick is when you apply it?}}\\\hdashline
    seller& \textbf{\textit{Yes, exactly. Glossy finishes are shinier, while matte finishes have no shine and cream finishes are somewhere in between.}}\\ \hdashline
    customer& I see, thank you for explaining that. I'm open to trying different finishes, so \textbf{no specific preference} there. \\\hdashline
    seller& You're welcome, that's great to know! Lipsticks can also have some \textbf{special features} like being \textbf{portable}, \textbf{long-lasting}, or \textbf{durable}. Is there anything among these that you think is important?\\ \hdashline
    customer& Portable, long-lasting and durable all sound nice but \textbf{not mandatory} for me. \\\hdashline
    seller& Good to know. The lipsticks are also categorized based on \textbf{skin type}. There's the \textbf{all}, \textbf{normal} and \textbf{dry} skin type. What's your skin type?\\ \hdashline
    customer& \textbf{I don't think that's something that would affect my choice, as I don't know my skin type.} \\\hdashline
    seller& No problem, we can skip that. Lastly, \textbf{customer review} and \textbf{price} are often considered. What are your thoughts on these aspects?\\ \hdashline
    customer& Well, \textbf{I don't really look at customer reviews and I'm flexible regarding the price}. \\\hdashline
    seller& It sounds like you're quite flexible, which makes it easier to find the right product for you. Given your preferences, might I suggest Maybelline New York Color Sensational Red Lipstick Matte Lipstick, Dynamite Red? This is based on your preference for dynamite red color and it's not liquid. The brand is well established and it's a fairly popular one. \\ \hdashline
    customer& That sounds perfect. Thank you for your help! \\
    \hline
    \end{tabular}
    \caption{A single-pass generation conversation from GPT-4. The dialogue plan is displayed at the top of the table, and the pre-determined customer preference is reflected in the conversation with \textbf{bold}, and the clarification question and answer are \textbf{\textit{italicized.}}} \label{tab:good_example}
\end{center}
\end{table*}

\begin{table*}[t]
\small
\begin{center}
    \begin{tabular}{l p{0.82\linewidth} }
    \hline
    \textbf{Product Category} & cable organizer bags \& cases \\ \hline
    \textbf{Wanted Features} & color: black \\ \hline
    \textbf{Optional Features} & form factor; size; price; material type \\ \hline
    \textbf{Unwanted Features} & brand: Sunsir \\ \hline 
    \end{tabular}
    \begin{tabular}{l p{0.9\linewidth} }
    \hline
    \textbf{Speaker} & \textbf{Utterance} \\
    \hline
    customer& Can you recommend a good cable organizer bag or case for me? I prefer one with a lot of compartments. \\ \hdashline
    seller& Sure, I'd be happy to assist you in finding the perfect cable organizer. Could you please let me know your preferred color for the organizer? Would you like a black, grey, or maybe even a pink one? And how about the form? Do you prefer it to be in the form of a bag, case, or a sleeve?\\ \hdashline
    customer& I'll go with black as my color preference. And what do you mean by form factor?\\ \hdashline
    seller& Sure, I understood your color preference. With regards to the form factor, it refers to the design and physical shape of the cable organizer. For example, some are designed like a bag with a zipper, some come in the form of a hard case, while others are like a sleeve. Could you please tell me about your brand preference? There are several well-known brands such as Bagsmart, Bubm, and Inateck.\\\hdashline
    customer& I don't have a specific brand preference, but I don't want Sunsir.\\\hdashline
    seller& I see, let's narrow down the options further. Regarding the size of the cable organizer, are you looking for a large one that can store many items, a small, compact one for easier carry, or a one-size fits all type of organizer?\\\hdashline
    customer& I'm flexible with the size.\\\hdashline
    seller& Great! Now, considering the material of the cable organizer, do you have a preference? Some people like nylon ones for the durability, while others may choose polycarbonate or plastic options for their lightweightness. Do any of these resonate with your needs?\\\hdashline
    customer& What do you mean by material type?\\\hdashline
    seller& By "material type", I meant the fabric or substance that the organizer is made from. Cable organizers can be made from a variety of materials. Here are a few examples:\\
    &1. **Nylon**: This is a type of synthetic fabric that's known for its strength and durability. It's also generally resistant to dirt and easy to clean.\\
    &2. **Polycarbonate or Plastic**: Some organizers are designed with a hard exterior for added protection. These types are lightweight, and plastic ones are particularly good for water resistance.\\
    &3. **Leather**: There are also organizers made from leather or PU leather. They offer a sleek, stylish look while also providing a good level of protection.\\
    &Each material comes with its own advantages. Nylon is usually longer-lasting, polycarbonate or plastic is lighter and waterproof, and leather gives a classy touch. Do any of these materials align with your preferences?\\\hdashline
    customer& I think I would like an organizer made of nylon for its durability. Is there any specific one you would recommend that meets my criteria? It should be black, not in the Sunsir brand, flexible in size, and made of nylon.\\\hdashline
    seller& Based on your preferences for a black, nylon-made cable organizer that's not from the Sunsir brand, I have a few suggestions for you:\\
    &1. The first is the [BAGSMART Universal Travel Cable Organizer](https://www.amazon.com/BAGSMART-Universal-Organizer-Electronics-Accessories/dp/B017SKRWL4). It's a great choice with several compartments to store your items such as a 9.7 inch iPad, Kindle, and Power Adapter. It's black with a hint of red color that gives it an edgy look.\\
    &2. The second option is [Inateck Electronics Organizer](https://www.amazon.com/Inateck-Electronics-Organizer-Splashproof-Waterproof/dp/B07ZV2CZSR). It's also a black, splashproof and waterproof travel cable organizer that should serve your needs well.\\
    &3. The third one is [Luxtude Electronics Organizer](https://www.amazon.com/Luxtude-Electronics-Organizer-Portable-Essentials/dp/B086M63FXV). \\
    &This Travel Cable Case is a portable tech bag perfect for cable storage, cord storage and storing electronic accessories/USB/SD/Charger. It's also black, aligning with your color preference.\\
    &Please let me know if you need more information about these products or if there's anything else I can assist you with.\\ \hdashline
    customer& The BAGSMART Universal Travel Cable Organizer sounds like a good fit for me. I like the number of compartments it has and the aesthetic design. I'll go ahead and add it to my cart. Thank you very much for your assistance!\\
    \hline
    \end{tabular}
    \caption{Bad conciseness example by GPT-4 interactive generation. The seller's utterances are verbose.} \label{tab:bad_conciseness}
\end{center}
\end{table*}

\begin{table*}[t]
\small
\begin{center}
    \begin{tabular}{l p{0.82\linewidth} }
    \hline
    \textbf{Product Category} & fitted sheets \\ \hline
    \textbf{Wanted Features} & included components:1 fitted sheet; color: natural; number of pieces: 1 \\ \hline
    \textbf{Optional Features} & material; brand; fabric weight; size \\ \hline
    \textbf{Unwanted Features} & None \\ \hline 
    \end{tabular}
    \begin{tabular}{l p{0.9\linewidth} }
    \hline
    \textbf{Speaker} & \textbf{Utterance} \\
    \hline
customer& Hello there, I am interested in buying fitted sheets. Could you please recommend some for me?\\ \hdashline
seller& Could you please specify the size of the sheets you're looking for? For instance, do you need queen, twin, or full size? Also, could you share your preference for the material - would you prefer microfiber, cotton, or other fabric types?\\\hdashline
customer& I'd like one set of fitted sheets. As for the size and material, I don't have any particular preferences. However, could you please explain what you mean by "microfiber" material? I am not very familiar with this term.\\\hdashline
seller & Certainly! Microfiber is a type of synthetic material known for its softness, durability, and resistance to wrinkles and stains. \textbf{For the fitted sheets, would you prefer a pack with multiple pieces or just one sheet?}\\\hdashline
customer& Just one sheet is enough for me.\\\hdashline
seller& Given that you are looking for just one fitted sheet, would you have any color preferences or special design requirements for the sheet? Also, do you have any specific washing instructions in mind, such as machine wash or hand wash for your sheets?\\\hdashline
customer& As for the color, I would prefer something dark, maybe black or navy. As for washing, I think being machine washable would be most convenient for me.\\\hdashline
seller& I see that you have an inclination towards dark colors and machine washable fitted sheets. Moving forward, are you interested in any particular brand like Bare Home, Elegant Comfort, or Ambesonne? Also, do you have any preference for the weight of the fabric? For example, would you prefer 130, 110, or 100 grams per square meter fabric weight?\\\hdashline
customer& I don't really care about the brand or the weight of the fabric. I just need it to include one fitted sheet. Can you help me understand what "fabric weight" means?\\\hdashline
seller& Sure, I'd be happy to explain. The fabric weight is usually measured in grams per square meter (GSM). It's an indicator of the fabric's thickness and heaviness. A higher GSM means a thicker and heavier fabric. \textbf{Now, you mentioned that you only need one fitted sheet.} So, in your case, we want to ensure that the set you buy includes only one fitted sheet, instead of multiple pieces or additional items like pillowcases. Does that sound about right to you? Would you prefer a set that exclusively includes one fitted sheet, without any additional components?\\\hdashline
customer& Yes, that's right. I only need a set that includes one fitted sheet without any additional components.\\\hdashline
seller& I see that you prefer dark colors. However, I'd like to clarify if you'd only like black or navy, or if you're open to other dark colors like grey or dark blue as well? Also, would you be interested in a white color in case we don't have the dark color you prefer in stock?\\\hdashline
customer& I am open to other dark colors. However, I don't want a white one. I prefer the sheet to have a more natural look.\\\hdashline
seller& Great, I understand that you are looking for a dark-colored, machine-washable fitted sheet with a natural look. Taking into account your preferences, I'd like to recommend the BIOWEAVES 100\% Organic Cotton Twin Fitted Sheet. This sheet is made from 100\% organic cotton with a 300 thread count. It has a sateen weave, which gives it a soft, luxurious feel. The sheet comes in an undyed natural color which should fit your description of a 'natural look'. It is also GOTS certified and comes with a deep pocket design. Furthermore, it's machine washable for easy care. Priced at \$32.99, it's a great value for the quality. Would you like to proceed with this one?\\\hdashline
customer& Sure, that sounds perfect. I am happy with the BIOWEAVES 100\% Organic Cotton Twin Fitted Sheet you recommended. But I am not certain about what 'GOTS certified' and 'sateen weave' mean. Could you explain these terms before I make a purchase?\\
    \hline
    \end{tabular}
    \caption{Bad coherence and conciseness example by GPT-4 interactive generation. Bolded sentences are those the seller does not give the customer a second chance to make a value selection response and directly jumps into asking about new product aspects.} \label{tab:bad_coherence}
\end{center}
\end{table*}

\begin{table*}[t]
\small
\begin{center}
    \begin{tabular}{l p{0.82\linewidth} }
    \hline
    \textbf{Product Category} & countertop blenders \\ \hline
    \textbf{Wanted Features} & price: between \$50 and \$100; is dishwasher safe: yes; blade material: stainless steel \\ \hline
    \textbf{Optional Features} & country of origin; power source; number of speeds; department; color; voltage; style; wattage; included components; customer review; capacity\\ \hline
    \textbf{Unwanted Features} & brand: Kacsoo \\ \hline 
    \end{tabular}
    \begin{tabular}{l p{0.9\linewidth} }
    \hline
    \textbf{Speaker} & \textbf{Utterance} \\
    \hline

customer& Hi, I'm looking to purchase a countertop blender. Can you help me out?\\\hdashline
seller& Of course! Let's start with a couple of general features. Could you specify the price range you're hoping for the blender?\\\hdashline
customer& Yes, I'd like it to be between $50 and $100.\\\hdashline
seller& Great! Now, blenders also come with different types of blades. We usually recommend stainless steel as they're the most durable. Would you like a blender with a stainless steel blade?\\\hdashline
customer& Yes, that works for me. I want my blender to have a stainless steel blade.\\\hdashline
seller& I see. There are many brands that offer these features. However, I need to confirm, would you prefer a Kacsoo blender?\\\hdashline
customer& No, not really. I'd like to consider other brands instead.\\\hdashline
seller& No problem, we have plenty of other brands to choose from. Now, did you have a preference for the country of origin for the blenders? We have blenders made in Mexico, the USA, and China.\\\hdashline
customer& Not really, the country of origin is optional for me.\\\hdashline
 seller& I understand. Let's move on to the power source. Would you like a blender that's powered by \textbf{AC, electricity, or corded electricity}?\\\hdashline
 customer& That doesn't really matter to me. It's also optional.\\\hdashline
 seller& Fair enough. And what about the number of speed settings? We have blenders with up to 16 speed settings, as well as those with fewer options.\\\hdashline
customer& That's also optional for me.\\\hdashline
seller& Alright, I understand. How about the \textbf{department} and color? Some customers prefer their blenders match their kitchen's aesthetics.\\\hdashline
customer& Both the department and the color are optional to me.\\\hdashline
seller& Got it. Moving on, the voltage and style can also vary among blenders. Do you have a preferred \textbf{voltage} and style?\\\hdashline
customer& Those aspects are optional for me.\\\hdashline
seller& Understood. Now, what about the wattage and the components included with your blender?\\\hdashline
customer& Those aspects aren't really important to me. They are also optional.\\\hdashline
seller& Noted. And lastly, how about customer reviews and capacity? Some people prefer larger blenders and highly-rated products.\\\hdashline
customer& Those are optional aspects for me as well.\\\hdashline
seller& Thank you for all this information. Our blenders also come with different dishwasher safety levels. Is a dishwasher safe blender important to you?\\\hdashline
customer& Yes, I'd like my blender to be dishwasher safe.\\\hdashline
seller& Alright, considering all your preferences, I would like to recommend the "Oster 6812-001 Core 16-Speed Blender with Glass Jar, Black". It's priced at \$59.68, has a durable stainless steel blade, and also dishwasher safe. The other aspects that you mentioned as optional all meet your standards as well.\\\hdashline
customer& That sounds perfect. Thank you for your help.\\
    \hline
    \end{tabular}
    \caption{Bad feature example by GPT-4 single-pass generation. The inappropriately selected product aspects in bold make the conversation awkward.} \label{tab:bad_feature}
\end{center}
\end{table*}

\begin{table*}[t]
\small
\begin{center}

    \begin{tabular}{l p{0.82\linewidth} }
    \hline
    \textbf{Product Category} & cooling pads \\ \hline
    \textbf{Wanted Features} & manufacturer: Kootek \\ \hline
    \textbf{Optional Features} & None\\ \hline
    \textbf{Unwanted Features} & color: white \\ \hline 
    \end{tabular}

    \begin{tabular}{l p{0.9\linewidth} }
    \hline
    \textbf{Speaker} & \textbf{Utterance} \\
    \hline
customer& Hi, I'm looking for a cooling pad for my laptop.\\\hdashline
seller& Great! Let's start by narrowing down the options. Can you tell me what kind of manufacturer are you interested in?\\\hdashline
customer& Umm...I'm not sure. I've heard of Kootek, is that a good brand?\\\hdashline
seller& Absolutely! Kootek is one of our best-selling brands for cooling pads. Let's focus on their products for now. How about the color? Do you have a preference?\\\hdashline
customer & Ah, well...I don't want white. Is that an option?\\\hdashline
seller& Yes, we have cooling pads in different colors. How about red or blue?\\\hdashline
customer & Hmm...I think I like the sound of \textbf{blue}.\\\hdashline
seller & Great! Now, let's talk about \textbf{size}. What size is your laptop?\\\hdashline
customer & It's a \textbf{15-inch laptop}.\\\hdashline
seller & Perfect! We have a Kootek cooling pad that fits 12-17 inch laptops. Would you like to have \textbf{adjustable mounts for the laptop stand height angle}?\\\hdashline
customer & Sure, that sounds useful.\\\hdashline
seller & Great! Do you need any extra features like \textbf{USB ports or LED lights}?\\\hdashline
customer & LED lights would be cool, but I'm not sure about USB ports. Can you explain what they do?\\\hdashline
seller & Of course! USB ports allow you to connect other devices to your laptop while it's on the cooling pad. For example, you could plug in a mouse or an external hard drive.\\\hdashline
customer & Oh, that's handy. Yeah, I think I want USB ports then.\\\hdashline
seller & Great! Lastly, would you like a chill mat that has a \textbf{quiet fan or a more powerful fan}? Or would you prefer a balance between the two?\\\hdashline
customer & Hmm...I don't know much about fans. Can you explain the difference?\\\hdashline
seller & Sure thing! A quieter fan will provide a gentle breeze and might be better suited for everyday use, while a more powerful fan will provide a stronger airflow and might be better for gaming or heavy video editing.\\\hdashline
customer & Oh, I see. I think I'll go with the \textbf{balance between the two} then.\\\hdashline
seller & Excellent choice! Based on your preferences, I recommend our Kootek Laptop Cooling Pad 12"-17" Cooler Pad Chill Mat 5 Quiet Fans LED Lights and 2 USB 2.0 Ports Adjustable Mounts Laptop Stand Height Angle, Blue. It fits your budget and has all the features you're looking for. Would you like to take a look at it?\\\hdashline
customer & Yes, that sounds great. Can I get a link to the product page?\\\hdashline
seller & Of course! Here's the link. Let me know if you have any other questions or need further assistance.\\\hdashline
customer & Thanks for your help! I think I'm good to go.\\\hdashline
seller & You're welcome! Enjoy your shopping experience, and don't hesitate to reach out if you need anything else! \\
    \hline
    \end{tabular}
    \caption{Not instruction-following example by LLaMA2-70B-Chat single-pass generation. The only assigned required feature is the manufacturer of Kootek, and the only unwanted feature is white color.} \label{tab:not_instruction_following}
\end{center}
\end{table*}

\end{document}